\documentclass[lettersize,journal]{IEEEtran}
\IEEEoverridecommandlockouts                  
\usepackage{graphics} 
\usepackage{booktabs}
\usepackage{threeparttable}
\usepackage{stfloats}
\usepackage{makecell}
\usepackage{multirow} 
\usepackage{geometry}
\usepackage{array}  
\usepackage{amsmath}  
\usepackage{amssymb}  
\usepackage[export]{adjustbox} 
\usepackage{bm}
\usepackage{balance}
\usepackage{subcaption} 
\usepackage[font=small,labelfont=bf,tableposition=top,hypcap=false]{caption}
\usepackage{tikz}
\usepackage[compact]{titlesec}
\usepackage{algorithm, algorithmic}
\usepackage{cuted}    

\usepackage[pdftex,
        colorlinks=true,
        bookmarks=true,
        citecolor=red,
        linkcolor=blue,
        pagebackref=true,
        pdfstartview=Fit,
        breaklinks=true
            ]{hyperref}
\usepackage{cleveref}

\title{\LARGE\bf
MRCT: An Object-Centric Two-Layer Control Architecture for Heavy-Payload Multi-Robot Transportation
}
 \author{Ning Wang$^{1}$, Yoshihiko Nakamura$^{2}$, Tianwei Zhang$^{1,3*}$
 \thanks{$^{1}$The Shenzhen Institute of Artificial Intelligence and Robotics for Society}
\thanks{$^{2}$Mohamed bin Zayed University of Artificial Intelligence, Abu Dhabi, UAE}
 \thanks{$^{3}$The Chinese University of Hong Kong - Shenzhen, China}
\thanks{* Corresponding Author: {zhangtianwei@cuhk.edu.cn}}
}
\begin{document}
\maketitle

\titlespacing{\subsection}{0pt}{1.2ex}{0.6ex}

\begin{abstract}
Cooperative transportation of heavy payloads by heterogeneous mobile manipulators requires coordinated payload motion, force distribution, and compliant physical
interaction. This paper presents MRCT, an object-centric two-layer control architecture for multi-robot transportation. At the object level, a model
predictive controller (MPC) generates a coordinated contact-force reference for the payload. 
At the robot level, whole-body impedance control
(WBIC) modules on individual robots realise the reference through a confidence-based blending of velocity-tracking and impedance torques. 
The blending mechanism emphasises tracking when the joint-velocity error is small and increases compliance when tracking degrades because of abrupt reference changes or physical disturbances. 
Simulation and hardware experiments with a heterogeneous team comprising a single-arm and a dual-arm mobile manipulator transporting a near-rated table payload 
demonstrate accurate object tracking, regulated contact forces, and stable payload orientation.
Parameter sensitivity studies, obstacle-avoidance transportation, and step-load disturbance experiments further demonstrate the effectiveness of MRCT for reliable cooperative transportation under the evaluated conditions.
\end{abstract}
\section{Introduction}\label{char1}
Cooperative transportation by multiple robots enables the handling of heavy or bulky objects beyond the capability of a single platform. This capability is important in logistics, manufacturing, and service robotics, where payload motion, contact forces, and robot configurations must be coordinated under actuation and safety constraints. Heavy-payload transportation is particularly
challenging because the controller must regulate object motion, distribute interaction forces among robots with different capabilities, and remain effective under modelling errors, contact uncertainty, and external
disturbances~\cite{xiongrong}.

A central challenge is balancing trajectory tracking with physical interaction safety. Robots must remain sufficiently compliant to prevent impulsive contact,
force amplification, and possible structural damage, while retaining enough stiffness to support the payload and preserve its pose during motion \cite{COM}. 
This trade-off becomes more difficult for heterogeneous mobile manipulators, whose different morphologies, actuation limits, and contact configurations complicate
whole-body motion coordination and force realisation. Practical transportation tasks further require operation under obstacle constraints, manual reference changes, and varying payload loads.

Existing work addresses cooperative transportation through task and motion planning \cite{nancy-TRO}, payload-aware trajectory optimisation \cite{Nmpc}, coupled object-level control \cite{hrc}, and distributed or leader--follower coordination strategies~\cite{ZHANG2026,gaohuijun2024}.
However, many approaches either treat the mobile base and manipulator as separate subsystems, rely on specialised coordination structures, or focus on motion coordination without explicitly integrating compliant whole-body execution. In parallel, impedance and admittance control have been widely used to improve safety in physical interaction, but their integration with payload-level planning for cooperative transportation by heterogeneous mobile manipulators remains insufficiently explored.

This paper presents \emph{MRCT}, an object-centric two-layer control architecture for heavy-payload multi-robot transportation. An object-level MPC first computes a 
contact-force reference whose resultant defines a payload wrench reference. The resulting contact-force references are sent to the corresponding robot controllers, which independently execute local WBIC modules.
A confidence-based blending mechanism uses the instantaneous joint-velocity tracking error to interpolate between tracking and impedance torques: it
prioritises precise motion execution when tracking is reliable, and increases compliance when abrupt references or physical disturbances cause larger tracking deviations. This design permits a common payload-level objective while
allowing heterogeneous robots to respond locally to contact conditions. Code will be released upon acceptance.
The contributions of this paper are as follows:
\begin{itemize}
    \item \textbf{An object-centric two-layer architecture}, termed MRCT, that combines
    object-level MPC planning with robot-level WBIC execution for heavy-payload transportation by heterogeneous mobile manipulators.

    \item \textbf{A confidence-based torque blending strategy} that integrates velocity-tracking and impedance torques, enabling a smooth trade-off
    between tracking accuracy and compliant interaction.

    \item \textbf{A systematic simulation study} of the fusion and impedance parameters,
    characterising their effects on object tracking and contact-force
    regulation.

    \item Hardware validation with a heterogeneous single-arm/dual-arm mobile manipulator team in joystick-guided transportation, obstacle avoidance,
    and step-load disturbance scenarios.
\end{itemize}
 
\begin{figure}[t]
\centering
\includegraphics[width=\columnwidth]{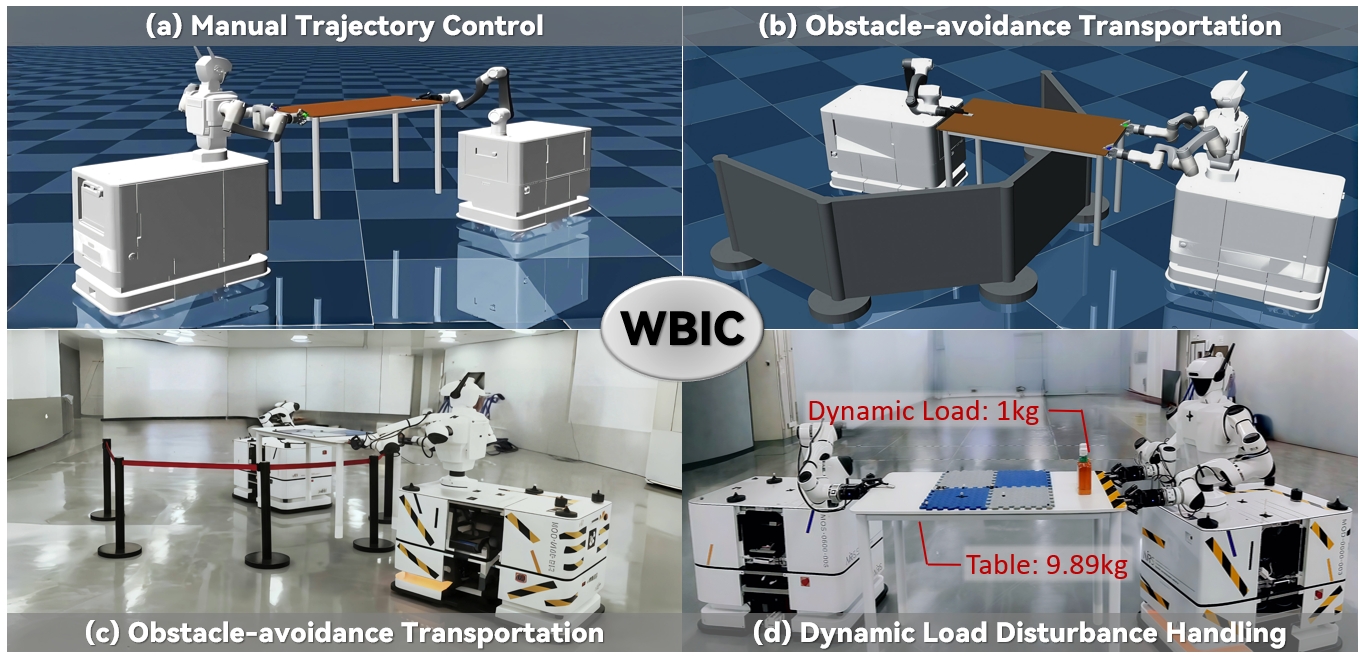}\\
\caption{A heterogeneous robot team comprising a single-arm and a dual-arm mobile manipulator collaboratively transports a \(9.89\,\mathrm{kg}\) table using the proposed MRCT framework. The task includes clockwise rotation around an obstacle and a \(1\,\mathrm{kg}\) step-load disturbance.}
\label{fig1.png}
\end{figure}

\section{Related Works}

Cooperative transportation has been studied through multi-robot task and motion planning, payload-aware trajectory optimisation, and coordinated manipulation.
For example, Motes et al.~\cite{nancy-TRO} formulate multi-robot task and motion planning using hypergraphs, while Kennel-Maushart and
Coros~\cite{kennel} consider payload-aware trajectory optimisation for non-holonomic mobile multi-robot manipulation with tip-over avoidance.
These methods provide valuable task-level planning capabilities, but often treat the mobile base and manipulators separately, which can limit the
whole-body coordination needed for force-sensitive heavy-payload transport.

More integrated cooperative control approaches employ decentralised MPC, leader--follower architectures, or capability-aware adaptation.
Gasbarrone et al.~\cite{Gasbarrone2024} develop a decentralised transportation method for humanoid robots using a leader--follower structure.
Lissandrini et al.~\cite{Nmpc} use decentralised nonlinear
MPC for heterogeneous aerial--ground cooperative manipulation, while Yan et al.~\cite{yanlei21} address capability differences through ability-aware adaptive control. Although effective in their respective
settings, these methods commonly target lightweight objects, assume specific coordination structures, or do not explicitly address compliant whole-body execution for heterogeneous mobile manipulators transporting heavy rigid
payloads.

Compliant interaction is often achieved using admittance or impedance control.
Admittance control adapts motion according to measured external forces but can be affected by feedback delays during fast contact transitions
\cite{stephane,keemink2018admittance}. Impedance control
instead specifies a force--motion relationship and is particularly suitable for fast local compliance in stiff physical interactions \cite{hogan1985impedance}. Whole-body impedance control extends this idea by coordinating full-body robot motion while regulating task-space compliance
\cite{Dietrich}. It has been applied successfully to legged and wheel-legged robots operating under uncertain contact conditions \cite{QuadrupedalLi,tengxun}.

\begin{figure*}[tbp]
\centering
\includegraphics[width=2\columnwidth]{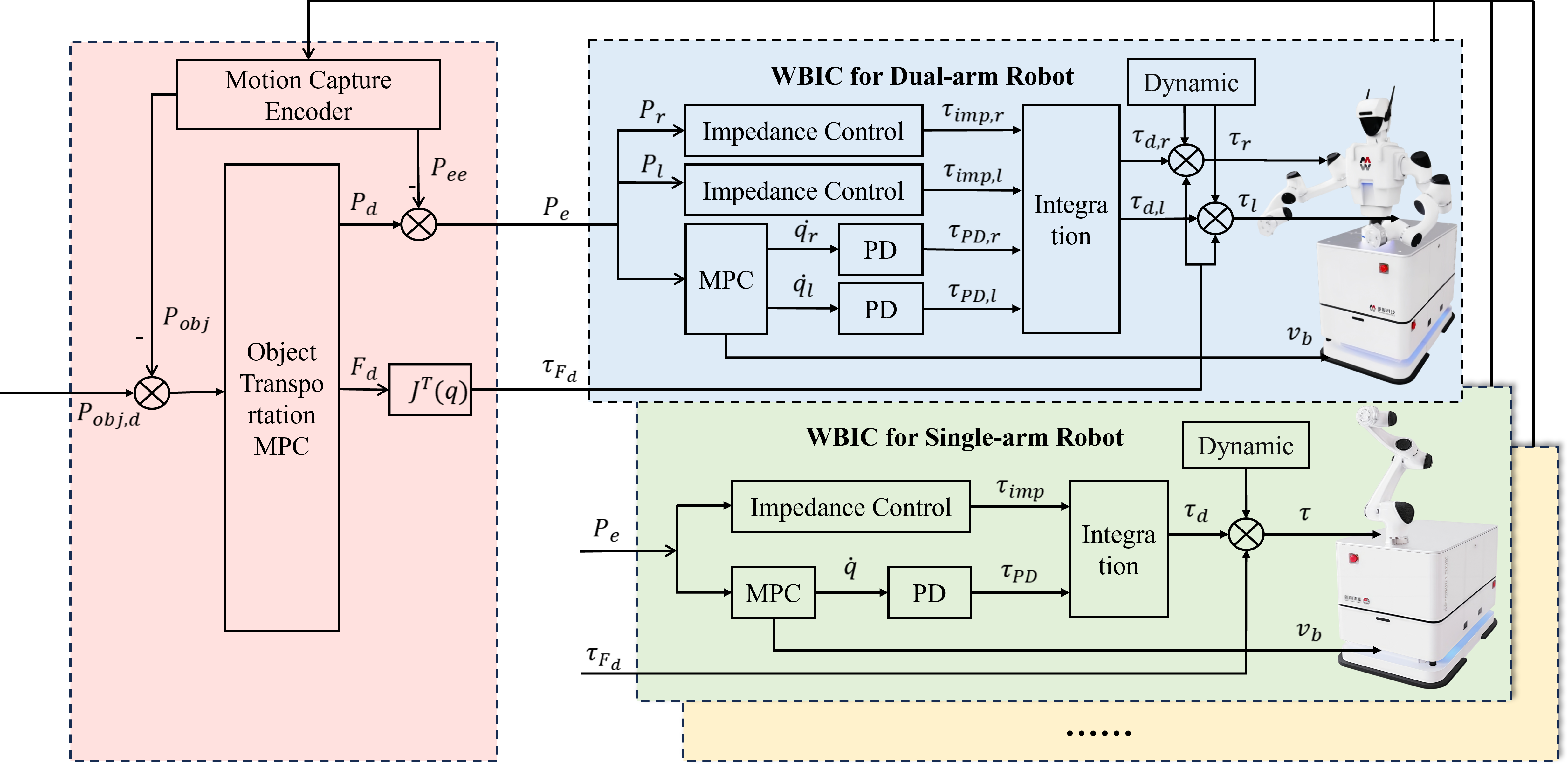}
\caption{Two-level MRCT framework. The object-level MPC computes coordinated contact-force references whose resultant defines the payload wrench reference.
Each robot executes a local WBIC module, with specialised configurations for the single-arm (green) and dual-arm (blue) robots.}
\label{fig:framework}
\vspace{-10pt}
\end{figure*}

In contrast to prior work, MRCT explicitly couples payload-level wrench planning with local whole-body compliant execution for a heterogeneous single-arm/dual-arm mobile-manipulator team. The proposed confidence-based blending mechanism provides an implementation-oriented interface between object-level planning and local contact control. Its effectiveness is evaluated through parameter sensitivity studies and real-world cooperative transportation tasks involving obstacles and step-load  disturbances.
\section{\textsc{Methods}}
Fig.~\ref{fig:framework} illustrates the proposed two-layer control framework for heterogeneous cooperative transportation.
The system operates on two levels. At the object level, a shared MPC module receives the object's current pose  \( P_{\text{obj}} \)  and desired trajectory  \( P_{\text{obj,d}} \).
It computes coordinated contact-force references and desired contact-point poses for the next control step.
This information is broadcast to all robots, providing a unified global objective.

At the robot level, each agent independently executes a WBIC scheme for the contact point(s) it manages. The robot first acquires its own state (base pose and joint angles), e.g., via SLAM, motion capture and encoders, then computes the current end-effector pose through forward kinematics. The resulting pose error is fed simultaneously to two parallel controllers. Specifically, a whole-body MPC generates desired joint velocities \( \dot{q} \)  and base velocity  \( v \). The base velocity \( v \) is sent directly to the base controller, while \( \dot{q} \) is passed to a joint-level PD controller to produce a velocity-tracking torque \( \tau_{\text{PD}} \). At the same time, an impedance controller uses the same pose error to compute a compliance torque  \( \tau_{\text{\text{imp}}} \).

These two torque components, \( \tau_{\text{PD}} \) and  \( \tau_{\text{\text{imp}}} \), are then integrated via a confidence-based weighting mechanism to yield a combined joint torque \( \tau_d \). The final joint torque command \( \tau \) is obtained by adding a feedforward term from the robot's dynamics and a compensation term \( \tau_{\text{fd}} \) for the object's transportation force.
This control logic is consistent across heterogeneous platforms. For a single-arm robot, one contact point is managed. For a dual-arm robot, the left and right arms are controlled independently by separate WBIC instances, generating distinct torque commands \( \tau_l \) and  \( \tau_r \).

\subsection{Object transportation MPC}
\label{object_mpc}
The proposed framework is hierarchically structured, beginning with an object-level MPC.  
It generates contact-force references whose resultant defines the net wrench applied to the payload. Let $N_c$ denote the number of active robot--object contact points. The spatial momentum dynamics of the object's centre of mass (CoM) are given by
\begin{equation}
\dot{h}_{\mathrm{com}} =
\begin{bmatrix}
\displaystyle\sum_{i=1}^{N_c} F_{d,i} + mg \\[2mm]
\displaystyle\sum_{i=1}^{N_c} r_i \times F_{d,i}
\end{bmatrix},
\label{eq:mpc}
\end{equation}
where $\dot{h}_{\mathrm{com}} \in \mathbb{R}^{6}$ denotes the rate of change of spatial momentum about the CoM, $m$ is the object mass, $g$ is the gravity vector, $F_{d,i} \in \mathbb{R}^{3}$ is the force applied at contact point $i$, and $r_i$ is the contact position relative to the CoM. In our experimental setup, $N_c=3$, corresponding to one contact on the single-arm robot and two contacts on the dual-arm robot.

An MPC is formulated to regulate the object's pose by solving a finite-horizon optimal control problem. The objective is to minimise the deviation from the desired trajectory while penalizing control effort, as defined by the cost function:
\begin{align}
\min_{\{u_t\}} \quad
& \sum_{t=0}^{N-1} \left( (P_{\text{obj},t} - P_{\text{obj,d},t})^\top Q (P_{\text{obj},t} - P_{\text{obj,d},t}) + u_t^\top R u_t \right) \notag \\
& + 
(P_{\mathrm{obj},N}-P_{\mathrm{obj},d,N})^\top
F(P_{\mathrm{obj},N}-P_{\mathrm{obj},d,N})
\label{eq:mpc_cost} \\
\text{s.t.} \quad
& X_{t+1} = A X_t + B u_t, \quad \forall t \label{eq:mpc_dynamics}  \\
& u_{\text{min}} \leq u_t \leq u_{\text{max}}, \quad \forall t \label{eq:mpc_force_constraint}
\end{align}

Here, the state is \(X_t = [P_{\text{obj},t}^\top, \dot P_{\text{obj},t}^\top, g^\top]^\top \in \mathbb{R}^{15}\), 
and the input
$u_t=[F_{d,1,t}^{\top},\ldots,F_{d,N_c,t}^{\top}]^\top$
collects the contact forces, and \(Q,R,F \succeq 0\) are standard weighting matrices. The matrices \(A, B\) result from linearising Eq. (\ref{eq:mpc}) around the nominal operating point. Constraints Eq.~\eqref{eq:mpc_dynamics} and ~\eqref{eq:mpc_force_constraint} enforce the linearised dynamics and box constraints on each contact force to ensure safe manipulation.

\subsection{Whole-body impedance control}
\label{sec:wbic}
The second layer of the architecture is a whole-body impedance control (WBIC) scheme executed independently on each robot. Its goal is to generate coordinated and compliant motion of the mobile base and manipulators so as to realise the object-level wrench reference.
\subsubsection{Whole-body MPC for motion generation}

To coordinate the motions of the mobile base and manipulators, a finite-horizon MPC is employed to generate desired motion references. Let $\mathcal{J}$ denote the set of manipulators on the robot, where $|\mathcal{J}|=1$ for the single-arm robot and $|\mathcal{J}|=2$ for the dual-arm robot. The controller computes optimal motions by minimizing end-effector tracking errors, base tracking errors, and control effort over the prediction horizon~\cite{SLQ-MPC,COM}:

\begin{equation}
\begin{aligned}
\min_{\{u_{t,j}\}} \quad
C ={}&
\sum_{t=0}^{T-1}
\Bigg[
\sum_{j\in\mathcal{J}}
\left(
\lVert x_{t,j}-x_{d,t,j}\rVert_{Q_j}^{2}
+
\lVert u_{t,j}\rVert_{R_j}^{2}
\right) \\
&\qquad\quad
+
\lVert x_{t,b}-x_{d,t,b}\rVert_{Q_b}^{2}
\Bigg] \\
&+
\sum_{j\in\mathcal{J}}
\lVert x_{T,j}-x_{d,T,j}\rVert_{F_j}^{2}.
\end{aligned}
\label{eq:wb_mpc_cost}
\end{equation}

Here, \(\|z\|_{Q}^{2}=z^\top Qz\), subject to robot kinematics, dynamics, and standard joint position and velocity constraints.
Solving the MPC yields the desired joint velocity $\dot q_d$ and base velocity $v_d$.

\subsubsection{Impedance control}
\label{sub.ic}

To ensure compliant interaction at the contact points, each manipulator simultaneously executes a Cartesian impedance controller. Let
$P_e = P_d - P$
denote the end-effector pose error. The desired interaction force is computed as~\cite{hogan1985impedance}

\begin{equation}
F_{\text{imp}} = D_{imp}\dot P_e + K_{imp} P_e
\label{eq:impedance}
\end{equation}
where $D_{imp}$ and $K_{imp}$ denote the virtual damping and stiffness matrices, respectively. The corresponding joint-space impedance torque is obtained through Jacobian transpose mapping:

\begin{equation}
\tau_{\text{imp}}
=
J^\top(q)\,F_{\text{imp}}
\label{eq:impedance_torque}
\end{equation}
 
This provides fast, local compliance to external disturbances and contact transitions.

\subsubsection{Confidence-based blending of tracking and impedance torques}
In cooperative transportation, the joint torque must handle both external disturbances, where impedance is desirable, and large tracking errors caused by abrupt trajectory changes.
Applying pure impedance or pure tracking in isolation can either amplify forces or compromise robustness. 
To balance these effects, a confidence-based blending mechanism is introduced.
Define the joint-velocity tracking error
\begin{equation}
e_v = \dot q_d - \dot q
\label{eq:ev}
\end{equation}
where $\dot q_d$ and $\dot q$ are the desired and actual joint velocities. A PD controller generates a velocity-tracking torque
\begin{equation}
\tau_{PD} = K_p e_v + K_d \dot e_v
\label{eq:tau_pd}
\end{equation}

with $K_p \succ 0$ and $K_d \succeq 0$ denoting proportional and derivative gain matrices. The blending weight $w \in [0,1]$ is chosen as
\begin{equation}
w = \frac{\sigma^2}{\|e_v\|^2 + \sigma^2}
\label{eq:weight}
\end{equation}
where $\sigma > 0$ sets the error scale of the transition between tracking-dominant and impedance-dominant behaviour.
When the tracking error is small, $w$ is close to $1$ and the controller emphasises precise velocity tracking; when the error is large, $w$ decreases and the controller relies more on impedance to remain compliant.

The final desired joint torque is computed as the weighted sum
\begin{equation}
\tau_d = w\, \tau_{\text{PD}} + (1-w)\, \tau_{\text{imp}}
\label{eq:tau_blend}
\end{equation}

The commanded joint torque is
\begin{equation}
\tau = \tau_d + \tau_{\mathrm{ff}} + \tau_{\mathrm{fd}},
\label{eq:commanded_torque}
\end{equation}
where $\tau_{\mathrm{ff}}$ denotes nominal robot-dynamics feedforward compensation and $\tau_{\mathrm{fd}}$ realises the contact-force reference generated by the object-level MPC. This formulation balances trajectory tracking and compliant interaction.
\begin{figure}[tbhp]
\centering
\includegraphics[width=\linewidth]{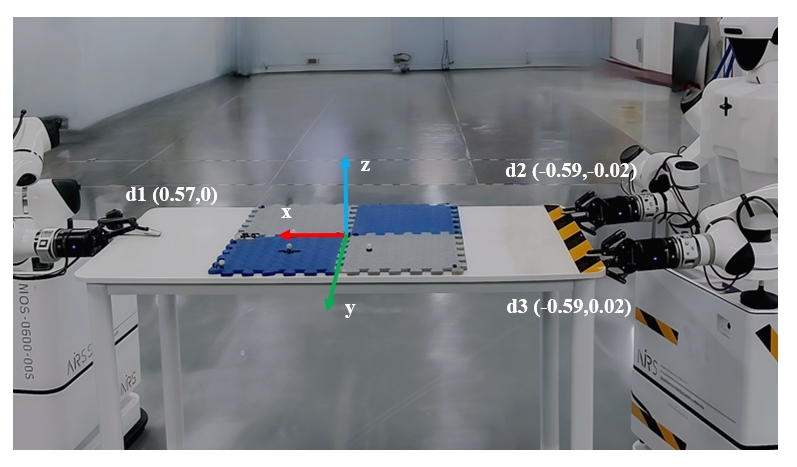}
\caption{Illustration of the grasping point settings.}
\label{fig:setup}
\end{figure}

\begin{figure*}[bhp]
\centering
\includegraphics[width=2\columnwidth]{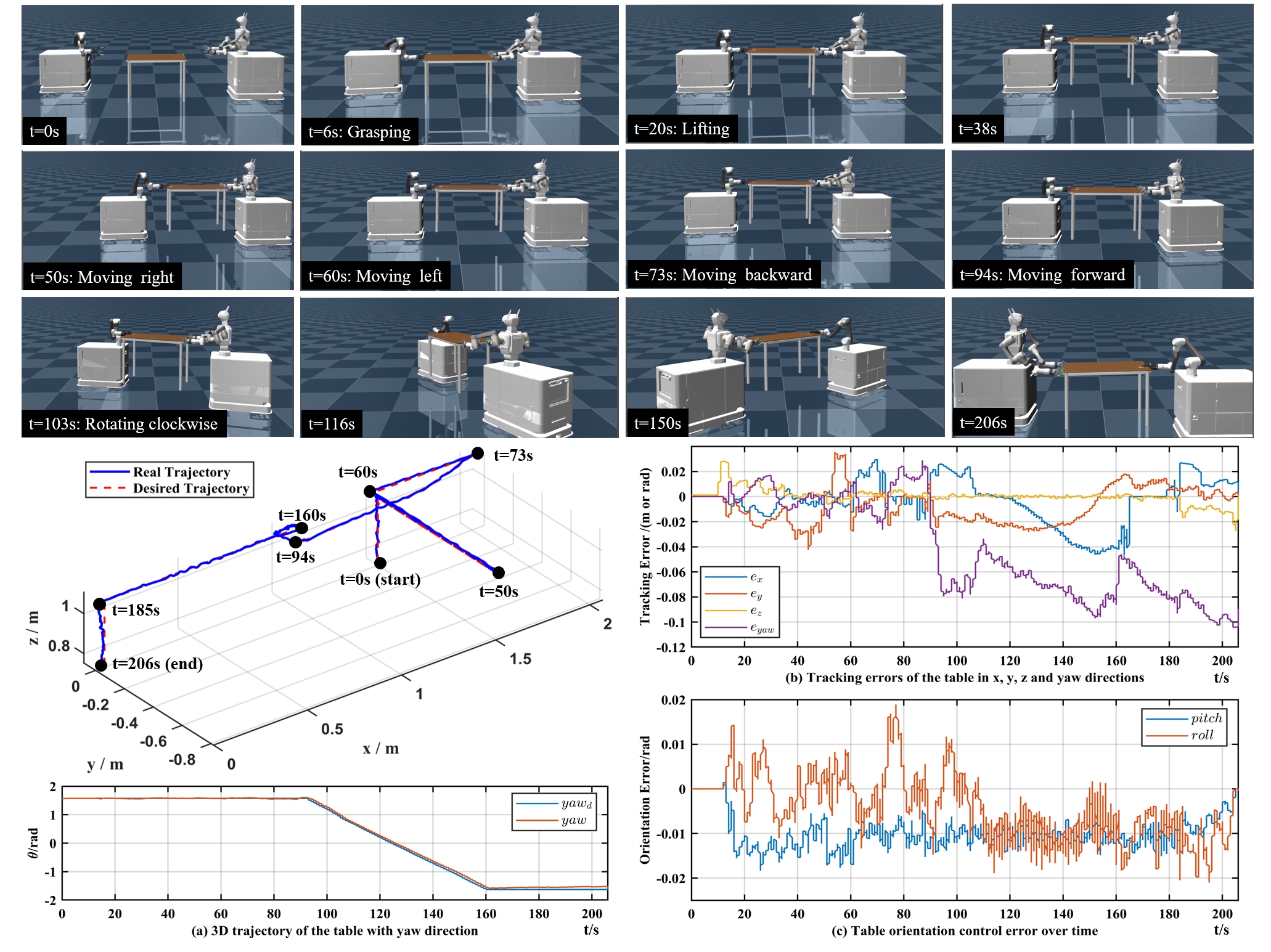}
\caption{
Simulation results for Experiment I using the proposed hybrid control. Top: Snapshots of key task phases. Bottom: (a) 3D trajectory and yaw orientation of the table; (b) tracking errors of the table along $x$, $y$, $z$, and yaw axes; (c) the $pitch$ and $roll$ angles of the table.
}
\label{fig:fangzhen}
\end{figure*}
\section{\textsc{Experimental Evaluations}}
This section evaluates the proposed control framework in simulation and on physical hardware. The hardware platform consists of a heterogeneous robot team comprising a single-arm and a dual-arm mobile manipulator. The single-arm robot is built on an AGV base equipped with an Elfin 5 arm (payload capacity 5\,kg), while the dual-arm robot uses an AGV base with two Elfin 3 arms (payload capacity 3\,kg per arm). The transported object is a rigid table of dimensions $1200\,\mathrm{mm} \times 600\,\mathrm{mm}$ and weight $9.89\,\mathrm{kg}$, and additional step-load  disturbances are introduced using two $500\,\mathrm{g}$ bottles placed on the tabletop. The grasping configurations are shown in Fig.~\ref{fig:setup}. Notably, neither robot can complete the transportation task independently; successful execution requires coordinated cooperative transportation under the proposed object-level MPC and WBIC architecture.





\begin{table*}[tbhp]
\centering
\small
\caption{Sensitivity to the blending-scale parameter \(\sigma\) in MuJoCo.}
\label{tab:exp1_wbc}
\begin{threeparttable}
{
\begin{tabular}{|c|c|c||c|c|c|c||c|c|c|}
\hline
\multicolumn{3}{|c||}{\textbf{Trial Info}} 
& \multicolumn{4}{c||}{\textbf{Position and Yaw Error (m/rad)}} 
& \multicolumn{3}{c|}{\textbf{Contact Force (N)}} \\
\hline
Trial & T (s) & $\sigma$
& $e_{x,\mathrm{pk}}$ / $\bar{e}_x$
& $e_{y,\mathrm{pk}}$ / $\bar{e}_y$
& $e_{z,\mathrm{pk}}$ / $\bar{e}_z$
& $e_{\mathrm{yaw},\mathrm{pk}}$ / $\bar{e}_{\mathrm{yaw}}$
& $F_{d1,\mathrm{pk}}$
& $F_{d2,\mathrm{pk}}$
& $F_{d3,\mathrm{pk}}$ \\
\hline
1 & 160 & 1.00 & 0.091 / -0.012 & 0.468 / -0.020 & 0.083 / -0.022 & 0.137 / -0.013 & 542.69 & 241.61 & 314.50 \\
2 & 147  & 0.75 & 0.092 /  0.009 & 0.233 / -0.016 & 0.154 / 0.045 & 0.145 / 0.033 & 191.51 & 128.32 & 115.64 \\
3 & 168  & 0.50 & 0.061 / -0.003 & 0.065 / 0.007 & 0.052 / 0.005 & 0.158 /  0.004 & 54.21 & 29.71 & 23.68 \\
4 & 180 & 0.25 & 0.205 / 0.018 & 0.102 / -0.013 & 0.054 / -0.021 & 0.120 / -0.019 & 48.65 & 20.51 & 24.13 \\
5 & Fail & 0.00 & - / - & - / - & - / - & - / - &- & - & - \\
\hline
\textbf{$|\cdot|_{\max}$} 
& \textbf{180} & --
& \textbf{0.373} / \textbf{0.020} 
& \textbf{0.468} / \textbf{-0.020}
& \textbf{0.154} / \textbf{0.045}
& \textbf{0.158} / \textbf{0.033}
& \textbf{542.69}
& \textbf{241.61}
& \textbf{314.50} \\
\hline
\end{tabular}
}
\end{threeparttable}
\end{table*}
\begin{table*}[tbp]
\centering
\small
\caption{Multiple trials in Experiment I under different impedance parameters}
\label{tab:exp1_hybrid}
\begin{threeparttable}
{
\begin{tabular}{|c|c|c|c||c|c|c|c||c|c|c|}
\hline
\multicolumn{4}{|c||}{\textbf{Trial Info}} 
& \multicolumn{4}{c||}{\textbf{Position and Yaw Error (m/rad)}} 
& \multicolumn{3}{c|}{\textbf{Contact Force (N)}} \\
\hline
Trial & T (s) & $K_{\text{imp}}$ & $D_{\text{imp}}$
& $e_{x,\mathrm{pk}}$ / $\bar{e}_x$
& $e_{y,\mathrm{pk}}$ / $\bar{e}_y$
& $e_{z,\mathrm{pk}}$ / $\bar{e}_z$
& $e_{\mathrm{yaw},\mathrm{pk}}$ / $\bar{e}_{\mathrm{yaw}}$
& $F_{d1,\mathrm{pk}}$
& $F_{d2,\mathrm{pk}}$
& $F_{d3,\mathrm{pk}}$ \\
\hline
1 & 206 & 100 & 100 & 0.138 / -0.021 & 0.140 / -0.005 & 0.048 / -0.003 & 0.105 / -0.038 & 52.69 & 24.61 & 38.50 \\
2 & 217 & 200 & 100 & 0.058 / -0.019 & 0.083 / -0.012 & 0.048 / 0.006 & 0.101 / -0.059 & 52.11 & 28.32 & 35.64 \\
3 & 248 & 300 & 100 & 0.103 / -0.035 & 0.115 / 0.017 & 0.052 / 0.005 & 0.118 / -0.004 & 54.21 & 29.71 & 23.68 \\
4 & 187 & 100 & 150 & 0.145 / -0.025 & 0.052 / 0.008 & 0.045 / -0.003 & 0.060 / -0.002 & 48.65 & 30.51 & 27.13 \\
5 & 214 & 100 & 50  & 0.070 / -0.020 & 0.136 / 0.019 & 0.047 / 0.009 & 0.074 / -0.011 & 50.44 & 37.32 & 24.55 \\
\hline
\textbf{$|\cdot|_{\max}$} 
& \textbf{248} & -- & --
& \textbf{0.145} / \textbf{-0.035}
& \textbf{0.140} / \textbf{0.019}
& \textbf{0.052} / \textbf{0.009}
& \textbf{0.118} / \textbf{-0.059}
& \textbf{54.21}
& \textbf{37.32}
& \textbf{38.50} \\
\hline
\end{tabular}
}
\end{threeparttable}
\end{table*}

To showcase the performance of the system in executing the cooperative transportation task, three main evaluation metrics are considered: position tracking accuracy, contact-force regulation, and payload orientation
stability.
 
\textbf{Experiment I: Manual Trajectory Control} Trajectories are generated via joystick input to evaluate system responsiveness and the performance under different control parameter settings.

\textbf{Experiment II: Obstacle-Avoidance Transportation} Transportation tasks are performed in the presence of predefined obstacles to evaluate contact force regulation and motion control accuracy.

\textbf{Experiment III: Step-Load Disturbance}
Two \(500\,\mathrm{g}\) bottles are placed on the table during transport, creating step changes in the payload load to evaluate system stability under disturbances.

\subsection{Result of Experiment I}
\label{exp1}
To analyze the impact of control parameters on system performance, sensitivity analysis experiments are conducted in the MuJoCo simulation environment. The study focuses on both the fusion weight parameter and the impedance control parameters by varying key configurations, and evaluates the resulting tracking errors and contact force responses under different conditions.

First, a sensitivity analysis is performed on the fusion weight parameter $\sigma$ in Eq.~\eqref{eq:weight}. The parameter $\sigma$ is defined as a proportion of the maximum joint velocity, and five different values are selected for evaluation. The corresponding results under different fusion settings are summarized in Table~\ref{tab:exp1_wbc}.

Table~\ref{tab:exp1_wbc} reports the sensitivity of MRCT to the blending scale $\sigma$. Small $\sigma$ values increase the impedance contribution, whereas larger values retain the tracking contribution over a broader error range. The results show a clear tracking--force trade-off, and the intermediate setting used in the subsequent experiments provides the most balanced performance.

Based on this, a sensitivity analysis is further conducted on the impedance model parameters in Eq.~\eqref{eq:impedance}. 
Multiple combinations of $(K_{\mathrm{imp}}, D_{\mathrm{imp}})$ are evaluated, and the resulting tracking errors and contact forces are summarised in Table~\ref{tab:exp1_hybrid}.


\begin{figure*}[tbhp]
\centering
\includegraphics[width=2\columnwidth]{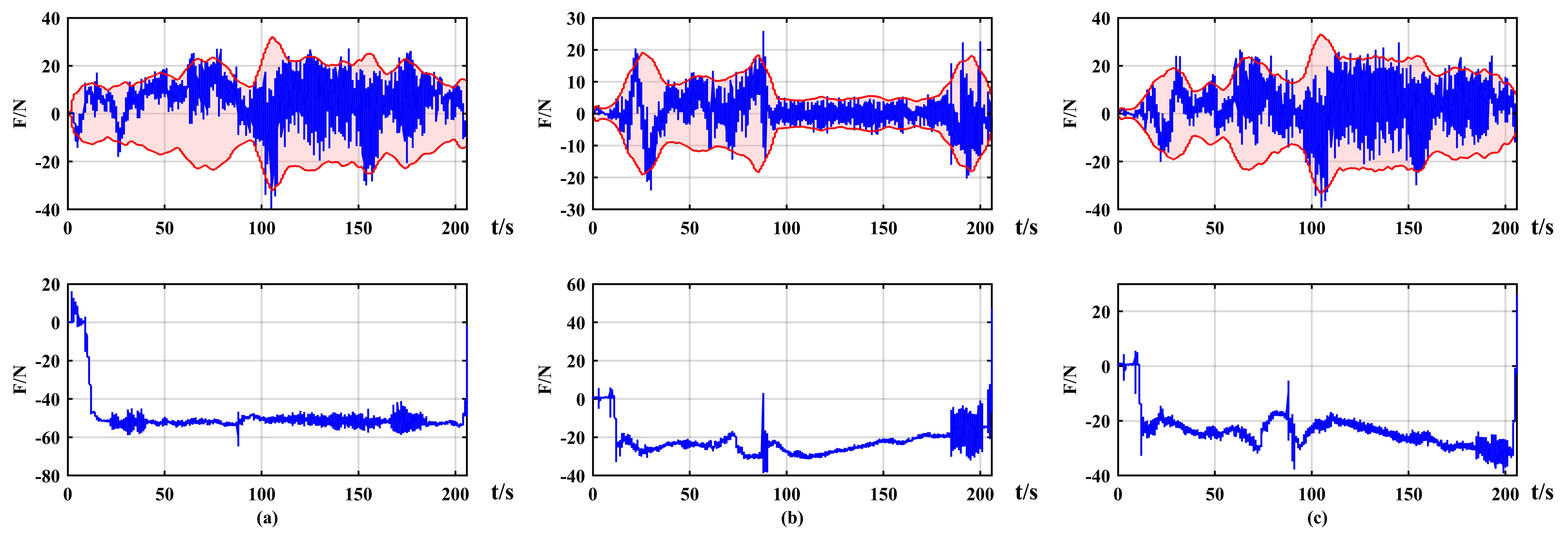}
\caption{
Contact force analysis from Experiment I. (a) Single-arm robot forces. (b) Right arm and (c) left arm forces of the dual-arm robot. Each top subfigure shows the measured contact force sequences (blue) between the three contact points and the table and their envelopes (red). 
}
\label{fig:fangzhenforce}
\end{figure*}

Fig.~\ref{fig:fangzhen} illustrates a representative hybrid trial. The robots track joystick commands to lift, translate, and rotate the table while maintaining smooth object motion without visible oscillations or slippage.
Position errors remain within $0.045\,\mathrm{m}$ in all Cartesian directions and yaw errors within $0.105\,\mathrm{rad}$, while pitch and roll stay within $\pm 0.018\,\mathrm{rad}$ and $\pm 0.021\,\mathrm{rad}$, respectively. 
The contact force profiles in Fig.~\ref{fig:fangzhenforce} show that forces remain within approximately $\pm 50\,\mathrm{N}$ for the single-arm robot and $\pm 25\,\mathrm{N}$ for the dual-arm robot during transportation, with the impedance controller dissipating transient peaks and keeping non-persistent spikes below $50\,\mathrm{N}$. 
These results indicate that the hybrid strategy achieves a favourable balance between trajectory accuracy and interaction safety, combining the tracking benefits of whole-body control with the force-limiting advantages of impedance.

\begin{figure*}[bhp]
\centering
\includegraphics[width=2\columnwidth]{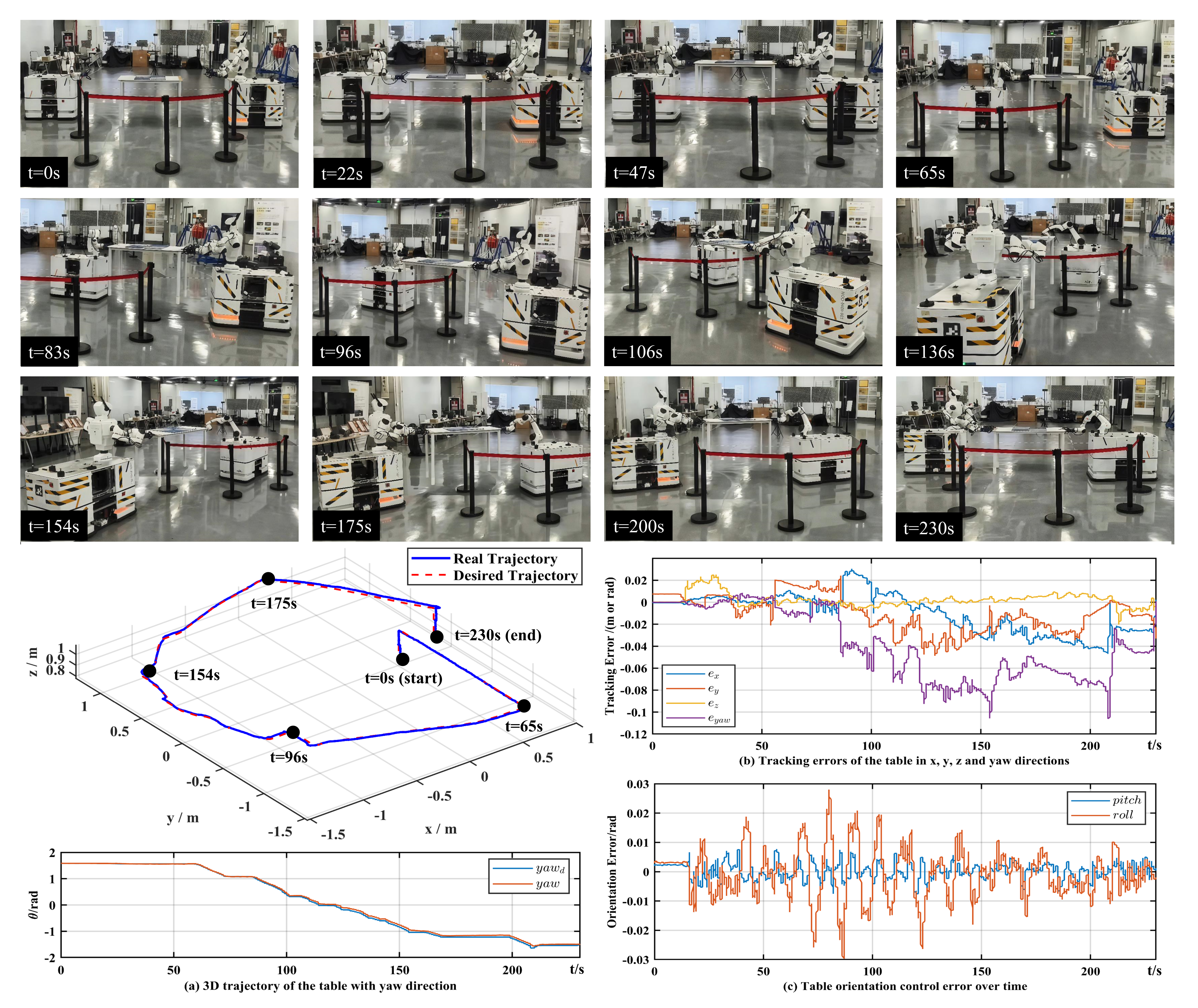}
\caption{Real-world results from Experiment II (obstacle-avoidance transportation). Top: Snapshots of the cooperative transportation process. Bottom: (a) 3D trajectory and yaw orientation; (b) tracking errors; and (c) pitch and roll angles of the table during obstacle traversal.}
\label{fig:shiwu}
\end{figure*}

\subsection{Result of Experiment II}
\label{exp2}
Experiment~II evaluates the proposed framework on physical hardware using a predefined obstacle-avoidance trajectory. Five real-world trials were conducted with the hybrid WBIC–MPC controller. As summarised in Table~\ref{tab:exp2}, the system maintains accurate position tracking and orientation stability across all trials, with performance metrics comparable to the simulation results of Experiment~I. The slightly improved tracking accuracy relative to the joystick-driven case is mainly due to the smooth, predefined
trajectory, which avoids the abrupt command changes inherent to manual operation.

Figure~\ref{fig:shiwu} shows a representative obstacle-avoidance trial. The robots approach, lift, transport, and return the table while following the desired path around four obstacles. Throughout the execution, positional errors remain within $0.045\,\mathrm{m}$ in all Cartesian directions and yaw errors do not exceed $0.105\,\mathrm{rad}$. The payload maintains excellent stability, with
pitch and roll angles constrained within $\pm 0.018\,\mathrm{rad}$ and $\pm 0.021\,\mathrm{rad}$, respectively. These results confirm that the proposed control strategy transfers reliably from simulation to real hardware and supports precise, stable cooperative transportation in cluttered environments.

\begin{table*}[tbp]
\centering
\small
\caption{Experiment II: cooperative transportation in a real obstacle-avoiding scene}
\label{tab:exp2}
{
\begin{threeparttable}
\begin{tabular}{|c|c|c|c||c|c|c|c||c|c|c|}
\hline
\multicolumn{4}{|c||}{\textbf{Trial Info}} 
& \multicolumn{4}{c||}{\textbf{Position and Yaw Error (m/rad)}} 
& \multicolumn{3}{c|}{\textbf{Contact Force (N)}} \\
\hline
Trial & T (s) & $K_{\text{imp}}$ & $D_{\text{imp}}$
& $e_{x,\mathrm{pk}}$ / $\bar{e}_x$
& $e_{y,\mathrm{pk}}$ / $\bar{e}_y$
& $e_{z,\mathrm{pk}}$ / $\bar{e}_z$
& $e_{\mathrm{yaw},\mathrm{pk}}$ / $\bar{e}_{\mathrm{yaw}}$
& $F_{d1,\mathrm{pk}}$
& $F_{d2,\mathrm{pk}}$
& $F_{d3,\mathrm{pk}}$ \\
\hline
1 & 230 & 100 & 100 & 0.049 / -0.006 & 0.043 / -0.006 & 0.038 / 0.006 & 0.135 / -0.015 & 47.57 & 27.64 & 36.35 \\
2 & 217 & 200 & 100 & 0.055 / -0.009 & 0.053 / -0.012 & 0.035 / 0.007 & 0.121 / 0.019 & 52.13 & 32.37 & 35.64 \\
3 & 180 & 300 & 100 & 0.103 / -0.005 & 0.049 / 0.007 & 0.032 / 0.005 & 0.058 / -0.027 & 54.21 & 29.73 & 32.81 \\
4 & 153 & 100 & 150 & 0.045 / 0.005 & 0.102 / 0.008 & 0.045 / 0.013 & 0.130 / 0.012 & 48.62 & 30.52 & 33.10 \\
5 & 216 & 100 & 50  & 0.072 / -0.018 & 0.066 / 0.019 & 0.057 / 0.016 & 0.074 / 0.011 & 60.40 & 37.20 & 34.58 \\
\hline
\textbf{$|\cdot|_{\max}$}
& \textbf{230} & -- & --
& \textbf{0.103} / \textbf{-0.018}
& \textbf{0.102} / \textbf{0.019}
& \textbf{0.057} / \textbf{0.016}
& \textbf{0.135} / \textbf{0.027}
& \textbf{60.40}
& \textbf{37.20}
& \textbf{36.35} \\
\hline
\end{tabular}
\end{threeparttable}
}
\end{table*}

\begin{figure*}[hbp]
\centering
\includegraphics[width=2\columnwidth]{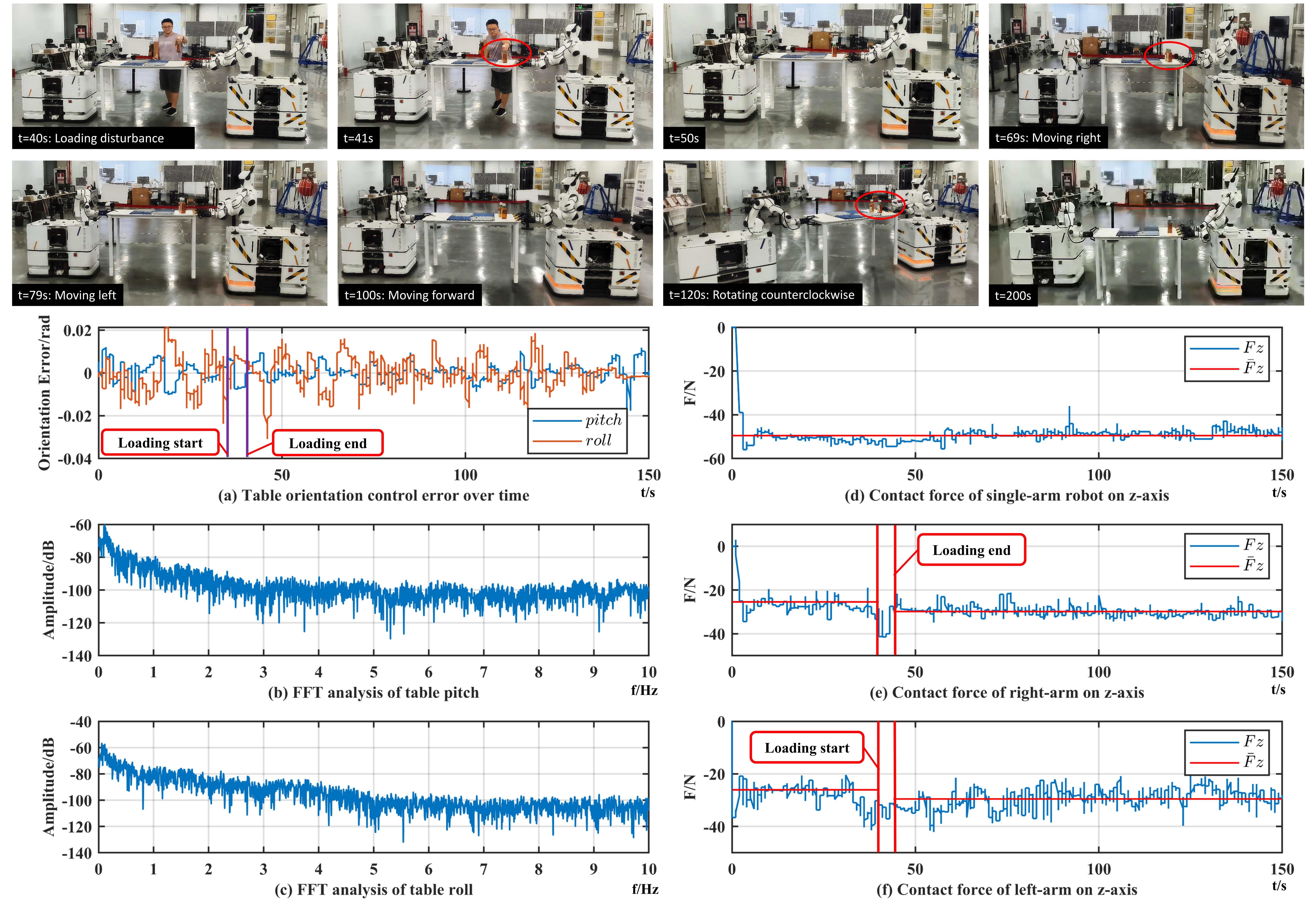}
\caption{A representative trial from Experiment III involving step-load  disturbance in the real-world. The top part shows selected visual snapshots illustrating key phases of the task execution. The bottom part illustrates the following: \textbf{(a)} the $pitch$ and $roll$ angles of the table; \textbf{(b)} and \textbf{(c)} the FFT analysis of the $pitch$ and $roll$ angles; \textbf{(d)}-\textbf{(f)} the contact forces along the $z$-axis measured by the single-arm robot, and the right and left arms of the dual-arm robot, respectively.
}
\label{jiashui}
\end{figure*}

\subsection{Result of Experiment III}\label{exp3}
To evaluate the robustness of the proposed framework, Experiment~III introduces step-load disturbances during cooperative transportation. Building on the joystick-based manual control setup of Experiment~I, two $500\,\mathrm{g}$ bottles are placed on the table surface near the dual-arm robot at different times during the task. As summarised in Table~\ref{tab:exp3}, the system maintains stable performance under these dynamic conditions: peak position tracking errors remain below $0.052\,\mathrm{m}$ with mean errors under $0.019\,\mathrm{m}$, while contact forces along the $x$-axis are bounded within $54.22\,\mathrm{N}$, $37.73\,\mathrm{N}$, and $39.81\,\mathrm{N}$ for the three contact points. The table’s orientation stability is preserved with pitch and roll peaks not exceeding $0.020\,\mathrm{rad}$ and $0.035\,\mathrm{rad}$, respectively, indicating effective disturbance rejection.

\begin{table*}[tbhp]
\centering
\small
\caption{Experiment III: cooperative transportation in the real-world}
\label{tab:exp3}
\begin{threeparttable}
\begin{tabular}{|c|c||c|c|c|c||c|c|c||c|}
\hline
\multicolumn{2}{|c||}{\textbf{Trial Info}} 
& \multicolumn{4}{c||}{\textbf{Position and Yaw Error (m/rad)}} 
& \multicolumn{3}{c||}{\textbf{Contact Force (N)}} 
& \multicolumn{1}{c|}{\textbf{Orientation (rad)}} \\
\hline
Trial & T (s) 
& $e_{x,\mathrm{pk}}$ / $\bar{e}_x$
& $e_{y,\mathrm{pk}}$ / $\bar{e}_y$
& $e_{z,\mathrm{pk}}$ / $\bar{e}_z$
& $e_{\mathrm{yaw},\mathrm{pk}}$ / $\bar{e}_{\mathrm{yaw}}$
& $F_{d1,\mathrm{pk}}$
& $F_{d2,\mathrm{pk}}$
& $F_{d3,\mathrm{pk}}$
& $pitch_\mathrm{pk}$ \quad $roll_\mathrm{pk}$ \\
\hline
1 & 150 & 0.038 / 0.004 & 0.042 / 0.011 & 0.042 / 0.002 & 0.050 / 0.011 & 51.73 & 35.61 & 32.53 & 0.017 \quad 0.032 \\
2 & 217 & 0.052 / 0.019 & 0.013 / 0.006 & 0.038 / 0.006 & 0.081 / 0.029 & 52.18 & 29.34 & 35.62 & 0.014 \quad 0.027 \\
3 & 148 & 0.043 / 0.005 & 0.049 / 0.007 & 0.012 / 0.005 & 0.018 / 0.007 & 54.22 & 37.73 & 39.81 & 0.020 \quad 0.035 \\
\hline
\textbf{$|\cdot|_{max}$}
& \textbf{217}
& \textbf{0.052} / \textbf{0.019}
& \textbf{0.049} / \textbf{0.011}
& \textbf{0.042} / \textbf{0.006}
& \textbf{0.081} / \textbf{0.029}
& \textbf{54.22}
& \textbf{37.73}
& \textbf{39.81}
& \textbf{0.020} \quad \textbf{0.035} \\
\hline
\end{tabular}
\end{threeparttable}
\vspace{-5pt}
\end{table*}

Fig.~\ref{jiashui} depicts a representative trial with step-load  disturbances. The table orientation (Fig.~\ref{jiashui}a) remains within approximately $0.032\,\mathrm{rad}$ in pitch and roll, consistent with the stability observed in Experiments~I and II. Frequency-domain analysis of these angles (Figs.~\ref{jiashui}b–c) shows that the pitch and roll spectra are dominated by low-frequency components, which is consistent with the smooth orientation response observed in the time domain.
The vertical contact forces (Figs.~\ref{jiashui}d–f) show the single-arm robot maintaining an average lifting force of about $-49.5\,\mathrm{N}$, while the dual-arm robot stabilises around $-29\,\mathrm{N}$ on each arm after bottle placement. The total vertical support force matches the combined weight of the table and bottles, confirming proper load distribution. Throughout the experiment, position tracking and orientation stability remain comparable to those of Experiments~I and II, and the step-load  stays securely positioned, demonstrating the robustness of the proposed collaborative transportation strategy under external disturbances.

Taken together, the experiments demonstrate that MRCT maintains accurate payload tracking and regulated contact forces across both simulation and hardware tests. 
The controller performs consistently under joystick-guided motion, obstacle-constrained transportation, and step-load disturbances, supporting its practical applicability to heterogeneous multi-robot transportation.

\section{Conclusion}
This paper introduced MRCT, an object-centric two-layer control architecture for heavy-payload cooperative transportation by heterogeneous mobile manipulators. MRCT combines object-level MPC planning with local WBIC
execution, where a confidence-based torque blending mechanism balances velocity tracking and compliant interaction. Simulation studies characterised the effects of the blending and impedance parameters on tracking and contact forces. Hardware experiments with a single-arm and a dual-arm mobile manipulator demonstrated accurate table transportation with regulated contact forces and stable orientation during joystick-guided motion, obstacle avoidance, and step-load disturbances.

\bibliographystyle{IEEEtran.bst}
\balance
\bibliography{IEEEabrv,bib}
\end{document}